\documentclass[sigconf]{acmart}

\usepackage{multirow}
\usepackage{amsmath}
\usepackage{booktabs}
\usepackage{tabularx}
\usepackage{array}

\AtBeginDocument{%
  }

\setcopyright{acmlicensed}
\copyrightyear{2026}
\acmYear{2026}
\acmDOI{XXXXXXX.XXXXXXX}
\acmConference[Conference acronym 'XX]{Make sure to enter the correct
  conference title from your rights confirmation email}{June 03--05,
  2026}{Woodstock, NY}
\acmISBN{978-1-4503-XXXX-X/2026/06}

\begin{document}

\title{Uncertainty-Aware Sea-Ice Type Mapping with
Multiple Ice Charts}

\author{Samira Alkaee Taleghan}
\affiliation{%
  \institution{University of Colorado Denver}
  \city{Denver}
  \state{Colorado}
  \country{USA}}
\email{samira.alkaeetaleghan@ucdenver.edu}

\author{Younghyun Koo}
\affiliation{%
  \institution{National Snow and Ice Data Center (NSIDC), CIRES, University of Colorado Boulder}
  \city{Boulder}
  \state{Colorado}
  \country{USA}}
\email{younghyun.koo@colorado.edu}

\author{Andrew P. Barrett}
\affiliation{%
  \institution{National Snow and Ice Data Center (NSIDC), CIRES, University of Colorado Boulder}
  \city{Boulder}
  \state{Colorado}
  \country{USA}}
\email{andrew.barrett@colorado.edu}

\author{Farnoush Banaei-Kashani}
\affiliation{%
  \institution{University of Colorado Denver}
  \city{Denver}
  \state{Colorado}
  \country{USA}}
\email{farnoush.banaei-kashani@ucdenver.edu}

\renewcommand{\shortauthors}{Alkaee Taleghan et al.}

\begin{abstract}
Sea-ice stage of development (SoD) describes the age and associated thickness of sea ice and provides important information for navigation, operational ice monitoring, and assessment of ice conditions. SoD labels are obtained from operational ice charts, where trained analysts interpret satellite observations and assign standardized stage codes to regions with similar ice conditions. These codes often represent ranges of compatible ice thicknesses rather than exact physical values.
Deep-learning methods can automate SoD mapping from satellite imagery and
commonly adopt operational ice charts as reference labels for training and
evaluation. These annotations are not exact, however; this is because chart interpretation relies on analyst judgement and on the observations available at the time, so different ice
services may assign different SoD labels to the same conditions. We term this
variation across independently produced expert annotations multi-annotator label
uncertainty; collapsing the annotations into a single deterministic target
discards this variation. A second source of uncertainty originates in the learned model
itself. In this paper, we quantify both sources: annotation uncertainty from disagreement among independent ice-service charts and model uncertainty from the learned predictive models. We then evaluate their relationship by testing whether model uncertainty is higher where ice services disagree. We observe that supervision incorporating information from multiple annotators can improve this correspondence, with soft supervision achieving the highest overall correlation of 0.256. The relationship becomes substantially stronger near the ice edge, where model predictive uncertainty closely tracks multi-annotator disagreement, reaching a correlation of 0.704 within 0--10 km. Among the uncertainty-estimation approaches, Monte Carlo dropout provides the best-calibrated confidence estimates, with an expected calibration error of 0.050.

\end{abstract}


\maketitle

\section{Introduction}

Sea ice plays an important role in the Arctic climate system and directly affects marine navigation and operations in polar regions \cite{Vihma}. Safe navigation requires timely information not only about ice presence and concentration but also about its physical characteristics. Sea-ice stage of development (SoD), which describes ice according to its age and associated thickness, is therefore an important parameter for operational monitoring and automated ice-chart production \cite{stokholm2024autoice,taleghan2024semisupervised}. Operational SoD information is produced by trained ice analysts, who interpret synthetic aperture radar (SAR) imagery together with other satellite, meteorological, and contextual information, delineate regions with similar ice conditions, and assign standardized ice information \cite{alkaeetaleghan2025icebench}. The World Meteorological Organization (WMO) defines the SoD classes through the WMO Sea-Ice Nomenclature \cite{wmo259}, while SIGRID-3 provides a standardized representation for operational ice charts \cite{sigrid3}. Ice regions are commonly described using the Egg Code, which reports total ice concentration (CT), partial concentrations of the three principal ice types (CA, CB, CC), and their corresponding stages of development (SA, SB, SC). These SoD categories generally represent thickness intervals rather than exact values; for example, thin first-year ice corresponds to approximately $30$--$70$~cm, medium first-year ice to $70$--$120$~cm, and thick first-year ice to thicknesses greater than approximately $120$~cm \cite{wmo259,sigrid3}.

Most supervised sea-ice mapping pipelines use operational ice charts as reference labels and optimize the model toward the charted class. Deep-learning methods provide an increasingly effective approach for automating this ice charting from satellite observations\cite{alkaeetaleghan2025icebench, Taleghan2025icefmbench,jalayer2025}. This formulation implicitly treats the assigned SoD label as ground truth. However, the annotation process contains multiple sources of uncertainty. First, multi-annotator label uncertainty arises because SoD annotations record
expert interpretation rather than direct physical measurement. Ambiguous SAR
signatures and changing surface and environmental conditions lead analysts to
resolve observations using contextual information and professional judgement, so
ice services charting independently may assign different SoD labels to the same
observation. Second, model uncertainty arises because the imagery may not
uniquely determine the SoD class and the learned parameters are themselves
uncertain. We characterize both and evaluate one against the other, quantifying
multi-annotator disagreement directly from co-charted scenes and testing whether
model uncertainty recovers it.

Treating uncertain annotations as exact targets can directly affect deep-learning training, particularly when experts assign different labels to comparable observations, because deterministic targets can turn legitimate interpretation differences into apparent label noise. The model is then forced to fit one interpretation, an average, or conflicting supervision, obscuring the distinction between prediction error and ambiguity in the reference data and potentially encouraging unjustified confidence. Uncertainty quantification in automated sea-ice mapping has focused mainly on predictive, epistemic, and aleatoric uncertainty \cite{piresdelima2023uncertainty,chen2023sicuncertainty,chen2023calibration,wulf2024panarctic,heffring2026bayesian}, while other studies document disagreement among ice analysts and services in region delineation, sea-ice concentration, and stage-of-development labels \cite{moen2013comparison,karvonen2015comparison,cheng2020agreement}. These findings show that uncertainty also exists in the expert annotations used for training, yet reference-label uncertainty—particularly for SoD—has received less attention, motivating methods that preserve it during model training.

This study focuses on uncertainty in operational SoD reference annotations and
how it should be handled during model training and uncertainty evaluation. We
treat each national ice service as an independent annotator: a chart is one
annotator's interpretation of a scene, and a scene charted by two or more
services carries that many independent annotations of the same observation. The
variation among those annotations is what we call multi-annotator label
uncertainty, and we quantify it per pixel as inter-annotator disagreement
$D(x)$, measured on the ordinal SoD ladder with the first Wasserstein distance
so that confusion between adjacent stages counts for less than confusion between
distant ones. We then compare five supervision strategies that retain or
collapse this information in different ways and evaluate model uncertainty in
four forms: predictive entropy, epistemic uncertainty from deep ensembles and
Monte Carlo dropout, an evidential Dirichlet head, and set-valued conformal
prediction. A control model trained to predict $D(x)$ directly bounds how much
of the disagreement is recoverable from the observations at all. This framework
connects the two sources, allowing their correspondence to be tested rather than
assumed.

The main contributions of this study are: (i) we construct a SoD dataset comprising 404 Sentinel-1 scenes independently charted by at least two of four national ice services (NIC, DMI, CIS, and NOAA); and (ii) we compare five supervision strategies that retain or collapse information from the independent annotations in different ways, and evaluate predictive entropy, ensemble-, dropout-, and evidential-based model uncertainty, conformal prediction sets, and a direct disagreement-regression control against variation measured directly from the expert charts rather than a single reference annotation.

The remainder of the paper is organized as follows. Section~2 reviews related work, Section~3 presents the methodology, Section~4 describes the experimental evaluation and results, and Section~5 concludes with the main findings and limitations.

\section{Related Work}
Deep-learning methods are widely used for automated sea-ice classification and ice-chart generation from SAR and multi-source observations. Previous work investigates supervised learning for sea-ice type and SoD classification \cite{taleghan2024semisupervised,alkaeetaleghan2025icebench}, and examines the behavior and interpretation of deep-learning models for operational ice charting \cite{jalayer2025}. However, most approaches still rely on operational ice charts as the primary source of reference labels, making the uncertainty associated with these annotations an important consideration for model training.
Uncertainty quantification receives increasing attention in automated sea-ice mapping. Pires de Lima and Karimzadeh \cite{piresdelima2023uncertainty} use model ensembles and dropout to estimate predictive uncertainty in sea-ice segmentation from Sentinel-1 SAR imagery. Chen et al.\ \cite{chen2023sicuncertainty} use a heteroscedastic Bayesian neural network to estimate epistemic and aleatoric uncertainty in sea-ice concentration retrieval, and subsequent work investigates calibration of these uncertainty estimates \cite{chen2023calibration}. Wulf et al.\ \cite{wulf2024panarctic} derive calibrated predictive uncertainty for pan-Arctic sea-ice concentration retrieval, while Heffring and Xu \cite{heffring2026bayesian} use a Bayesian Transformer to quantify uncertainty associated with model parameters and predictions. These studies demonstrate the importance of uncertainty estimation for reliable sea-ice products, but they primarily characterize uncertainty through the predictive model or its outputs.
A separate body of work shows that uncertainty also exists in the expert-generated ice charts that commonly serve as reference labels. Moen et al.\ \cite{moen2013comparison} compare independent ice charts produced from the same SAR observation and report substantial differences between analysts in both the delineation of ice regions and the assignment of stage-of-development labels. Karvonen et al.\ \cite{karvonen2015comparison} compare sea-ice concentration estimates from several groups of ice analysts and identify significant variation among expert estimates, particularly for intermediate ice concentrations. Cheng et al.\ \cite{cheng2020agreement} similarly quantify inter-analyst agreement at the Canadian Ice Service and show that estimates from individual analysts can differ even when they interpret the same SAR observations. The difficulty of interpretation also changes with ice conditions. During the melt season, changes in surface properties can make first-year ice and open water appear similar in SAR imagery, increasing ambiguity in the information available to analysts \cite{cheng2020agreement}. Thus, uncertainty does not arise only after a model produces a prediction; it already exists in the expert annotations used to train and evaluate the model. However, most uncertainty-aware sea-ice learning approaches do not explicitly preserve this annotation uncertainty when constructing the training target. This motivates our study of uncertainty directly in operational SoD labels and its effect on deep-learning models.

\section{Methodology}
\label{sec:methodology}

Our methodology separates uncertainty in the reference annotations from
uncertainty in the learned model. An individual operational ice chart can
already contain uncertainty because an SoD code can correspond to a range of
ice thicknesses or to several fine SoD classes. When several ice services
independently chart the same observation, their interpretations introduce an
additional source of uncertainty through disagreement among annotations. We
preserve these sources during target construction rather than assuming that
every chart provides a single exact ground-truth class. Separately, we estimate
predictive and epistemic uncertainty from the trained models and examine
whether these quantities correspond to uncertainty measured directly from the
expert annotations \cite{kendall2017uncertainties}.

\begin{table}[t]
\caption{SIGRID-3 stage codes and their thickness-based SoD representation.}
\label{tab:codes}
\centering
\footnotesize
\setlength{\tabcolsep}{3pt}
\renewcommand{\arraystretch}{1.08}
\begin{tabularx}{\columnwidth}{
    @{}
    l
    >{\raggedright\arraybackslash}X
    >{\raggedright\arraybackslash}p{0.25\columnwidth}
    r
    @{}
}
\toprule
Code & WMO stage & SoD representation & Thickness (cm) \\
\midrule
81, 82
    & New ice / nilas / ice rind
    & new
    & 0--10 \\

84
    & Grey ice
    & grey
    & 10--15 \\

85
    & Grey-white ice
    & grey-white
    & 15--30 \\

87, 88, 89
    & Thin first-year ice
    & thin FY
    & 30--70 \\

91
    & Medium first-year ice
    & medium FY
    & 70--120 \\

93
    & Thick first-year ice
    & thick FY
    & 120--200 \\

95, 96, 97
    & Old / second-year / multi-year ice
    & old
    & 200--300$^{\ddagger}$ \\

\midrule
83
    & Grey ice, undifferentiated
    & grey $\cup$ grey-white
    & 10--30 \\

86
    & First-year ice, undifferentiated
    & thin FY $\cup$ medium FY $\cup$ thick FY
    & 30--200 \\

\midrule
\multicolumn{4}{@{}p{\columnwidth}@{}}{
\scriptsize $^{\ddagger}$The WMO interval is open-ended; 300\,cm is used
to bound the ordinal thickness ladder.
} \\
\bottomrule
\end{tabularx}
\end{table}

\subsection{SoD Label Representation}
\label{sec:sod_representation}

Operational SoD labels originate from ice charts produced by trained ice
analysts. Each chart partitions the scene into polygons with similar ice
conditions and assigns each polygon a SIGRID-3 Egg Code. The Egg Code specifies
the total ice concentration $\mathrm{CT}$ and up to three principal ice types,
each described by a partial concentration ($\mathrm{CA}$, $\mathrm{CB}$, or
$\mathrm{CC}$) and a corresponding stage-of-development code
($\mathrm{SA}$, $\mathrm{SB}$, or $\mathrm{SC}$).
We harmonize these stage codes across ice services using their WMO thickness
intervals, as shown in Table~\ref{tab:codes}. The SoD task is conditioned on the presence of ice. We therefore remove the
open-water proportion and renormalize the remaining values over the ice
classes. These intervals are themselves
uncertain representations of ice thickness because a stage code specifies a
range of compatible thicknesses rather than an exact value. For example, thin
first-year ice represents approximately $30$--$70$~cm and medium first-year
ice $70$--$120$~cm. When a code spans several finer SoD classes, we retain the
full interval rather than assigning it to an arbitrary single class. For
example, code 86 represents undifferentiated first-year ice over approximately
$30$--$200$~cm and therefore includes thin, medium, and thick FY ice.

The Egg Code can report several ice types that coexist within the same polygon.
Their partial concentrations specify how much of the polygon is associated
with each type, but the chart does not specify the spatial location of those
types within the polygon. Therefore, every pixel inherits the complete
polygon-level SoD composition rather than a single locally identified ice
type. For ice service $a$ and pixel $x$, we represent this composition using
the reported partial concentrations as
\begin{equation}
\mathbf{q}_{a}(x)
\propto
\sum_{j\in\mathcal{J}_{a}(x)}
c_{a,j}(x)\,
\mathbf{e}\!\left(s_{a,j}(x)\right),
\label{eq:partial_mixture}
\end{equation}
where $j$ indexes the ice types reported in the Egg Code,
$c_{a,j}(x)$ is the corresponding partial concentration,
$s_{a,j}(x)$ is its SoD code, and $\mathbf{e}(s)$ converts that code to the
thickness-ordered SoD representation. The
resulting values are normalized to represent the relative SoD composition of
the ice within the polygon.

This representation describes the SoD information provided by a single ice
service. When different ice services independently chart the same satellite
observation, their corresponding compositions $\mathbf{q}_{a}(x)$ may also
differ because polygon delineation, ice composition, and stage assignment
depend on analyst interpretation. Thus, in addition to the thickness ranges
and polygon-level composition contained within an individual chart,
multi-service observations can exhibit disagreement between independently
produced SoD descriptions. The resulting $\mathbf{q}_{a}(x)$ preserves the thickness ranges and
ice composition reported by service $a$ at pixel $x$ and forms the basis for
the supervision strategies described in the following section.

\subsection{Uncertainty-Aware Supervision}
\label{sec:supervision}

Because each SoD label represents a thickness interval rather than an exact ice
thickness, uncertainty is already present within an individual label. Additional within-chart
uncertainty arises when a broader stage code spans several finer thickness
intervals or when multiple ice types with different partial concentrations are
reported within the same polygon. When multiple ice services independently
chart the same observation, their annotations can further differ because of
analyst interpretation, introducing multi-annotator disagreement.
When
multiple ice services chart the same observation, their independently produced
$\mathbf{q}_a(x)$ can also disagree. We evaluate supervision strategies that
retain or discard these two sources of information in different ways.

\textbf{Hard supervision.}
The conventional formulation reduces the annotation from one ice service to a
single dominant SoD class,
$y_a(x)=\arg\max_k q_{a,k}(x)$, and minimizes
$\mathcal{L}_{\mathrm{hard}}
=-\log p_{\theta}(y_a(x)\mid x)$.
This retains only the dominant stage and discards the remaining polygon
composition and any broader thickness range represented in
$\mathbf{q}_a(x)$. When several services are available, the hard target is
taken from one reference service using the fixed priority NIC, DMI, CIS, and
NOAA.

\textbf{Consensus supervision.}
For pixels charted by multiple services, let $\mathcal{A}(x)$ denote the set of
available services and $A_x=|\mathcal{A}(x)|$. We first average their SoD
compositions,

\[
\bar{\mathbf{q}}(x)
=
\frac{1}{A_x}
\sum_{a\in\mathcal{A}(x)}
\mathbf{q}_a(x),
\]

and define the consensus class as
$y_{\mathrm{cons}}(x)=\arg\max_k \bar q_k(x)$.
The corresponding loss is
$\mathcal{L}_{\mathrm{cons}}
=-\log p_{\theta}(y_{\mathrm{cons}}(x)\mid x)$.
Consensus therefore uses information from all available services to determine
a common dominant stage, but removes the remaining within-chart composition
and cross-service disagreement from the training target.

\textbf{Soft supervision.}
Instead of reducing the cross-service mean to one class, soft supervision
retains the complete mean SoD composition and minimizes

\[
\mathcal{L}_{\mathrm{soft}}
=
-\sum_{k=1}^{K}
\bar q_k(x)\log p_{\theta,k}(x).
\]

This preserves the relative proportions assigned to the different stages,
including both the composition reported within individual Egg Codes and
differences among services. This formulation is related to
distribution-valued supervision \cite{geng2016label}.

\textbf{Multi-annotator }
We quantify disagreement among the available ice-service annotations using the
first Wasserstein distance, which accounts for the ordering of the SoD
thickness intervals \cite{peyre2019computational}. For pixels with at least
two available services,

\begin{equation}
D(x)
=
\frac{2}{A_x(A_x-1)}
\sum_{\substack{a,b\in\mathcal{A}(x)\\a<b}}
W_1\!\left(\mathbf{q}_a(x),\mathbf{q}_b(x)\right).
\label{eq:disagreement}
\end{equation}

Because the classes are ordered by their associated thickness intervals,
disagreement between neighboring stages contributes less than disagreement
between stages farther apart. We use $D(x)$ as the continuous measure of
multi-annotator SoD disagreement throughout the experiments.

To test whether strongly contested annotations should contribute less to
training, we weight the soft loss according to
$w(x)=1/(1+D(x))$. The weighted objective is
$\mathcal{L}_{\mathrm{weighted}}
=w(x)\mathcal{L}_{\mathrm{soft}}$, with the pixel losses normalized by the sum
of the weights. Pixels on which the services agree therefore receive greater
weight than pixels with large disagreement.

Finally, we consider only the set of SoD stages supported by each service,
without requiring the model to reproduce the proportions within that set. Let

\[
\mathcal{S}_a(x)=\{k:q_{a,k}(x)>0\}
\]

denote the stages supported by service $a$. We define

\begin{equation}
\mathcal{L}_{\mathrm{AvgNLL}}
=
-\frac{1}{A_x}
\sum_{a\in\mathcal{A}(x)}
\log
\left(
\sum_{k\in\mathcal{S}_a(x)}
p_{\theta,k}(x)
\right).
\label{eq:avgnll}
\end{equation}
AvgNLL rewards probability assigned anywhere within the thickness range
supported by each service, without specifying how that probability should be
distributed within the range. It is therefore related to candidate-set or
partial-label learning \cite{cour2011learning}. Unlike soft supervision, it
retains the supported range but discards the partial-concentration proportions
within that range.

\subsection{Model Uncertainty}
\label{sec:model_uncertainty}

We evaluate model uncertainty separately from uncertainty encoded in the
reference annotations. Following the distinction between predictive and
epistemic uncertainty \cite{kendall2017uncertainties}, we first measure the
uncertainty of the predictive distribution and then estimate the component
associated with variation among model predictions.

\textbf{Epistemic uncertainty.}
For a predictive distribution $\mathbf{p}_{\theta}(x)$, predictive uncertainty
is measured using categorical entropy,
$H(x)=-\sum_{k=1}^{K}p_{\theta,k}(x)\log p_{\theta,k}(x)$
\cite{shannon1948}. We compare $H(x)$ with measured multi-annotator disagreement
using Spearman rank correlation $\rho(H,D)$. A positive correlation indicates
that observations assigned larger predictive uncertainty are also those for
which the ice services disagree more strongly.

We estimate epistemic uncertainty using deep ensembles
\cite{lakshminarayanan2017deep} and Monte Carlo dropout
\cite{gal2016dropout}. Given $M$ predictive distributions
$\{\mathbf{p}^{(m)}(x)\}_{m=1}^{M}$, their mean is
$\bar{\mathbf{p}}(x)=M^{-1}\sum_m\mathbf{p}^{(m)}(x)$. We estimate the
epistemic component as
\begin{equation}
U_{\mathrm{epi}}(x)
=
H\!\left(\bar{\mathbf{p}}(x)\right)
-
\frac{1}{M}
\sum_{m=1}^{M}
H\!\left(\mathbf{p}^{(m)}(x)\right).
\label{eq:epistemic}
\end{equation}
For deep ensembles, the predictions are obtained from independently trained
models. For Monte Carlo dropout, they are obtained from repeated stochastic
forward passes through the same network with dropout active at inference
\cite{gal2016dropout}. We compare $U_{\mathrm{epi}}(x)$ with $D(x)$ to test
whether epistemic uncertainty identifies observations on which independent ice
services disagree.

\textbf{Conformal prediction sets.}
We also evaluate set-valued uncertainty through split-conformal prediction
\cite{angelopoulos2023conformal}. Prediction sets $S(x)$ are calibrated on
validation pixels at miscoverage level $\alpha=0.1$ using the least-ambiguous
(LAC) score $1-p_{\theta,y}(x)$, together with the adaptive APS and RAPS
scores. Beyond verifying marginal coverage, we ask whether the sets are wide
in the right places: we correlate set size with $D(x)$, measure the coverage
gap between agreeing and contested pixels, and compare the prediction set with
the \emph{services' own answer set} $U(x)$ --- the union of the distinct modal
stages the available services assigned at $x$ --- using the rank correlation
$\rho(|S|,|U|)$ and the Jaccard overlap between the two sets. SIGRID-3 already
contains set-valued statements, so this comparison asks whether a calibrated
prediction set reproduces the ambiguity that analysts actually express.

\textbf{Disagreement learnability control.}
To assess whether multi-annotator disagreement is predictable from the available
observations, we train the same network backbone with a scalar output to
directly regress $D(x)$ using absolute error. This provides an empirical
reference for the amount of disagreement information recoverable from the SAR
and AMSR2 inputs. We additionally compare it with a constant predictor and
distance to the 15\% ice edge.
\section{Experimental Evaluation}

\subsection{Data}
\label{sec:data}

\begin{figure}[t]
\centering
\includegraphics[width=\columnwidth]{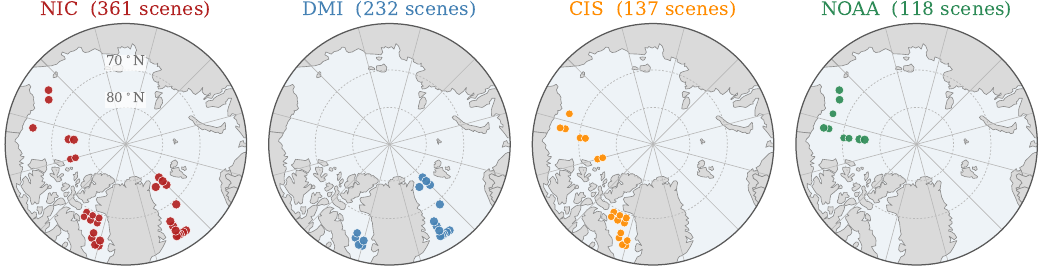}
\caption{Geographic coverage of operational ice charts in the multi-service dataset. Marker area indicates the number of Sentinel-1 scenes per spatial cell. NIC has the broadest coverage (361 scenes), followed by DMI (232), CIS (137), and NOAA (118).}
\label{fig:cov_map}
\end{figure}

We construct a multimodal sea-ice dataset for 2025 by pairing Sentinel-1 Synthetic
Aperture Radar (SAR) observations with operational ice charts produced by four
national ice services: the U.S. National Ice Center (NIC), the Danish
Meteorological Institute (DMI), the Canadian Ice Service (CIS), and the
U.S. National Oceanic and Atmospheric Administration (NOAA). Sentinel-1
Extra-Wide (EW) Ground-Range-Detected scenes are obtained from the Copernicus
archive and Google Earth Engine. We use the dual-polarization HH and HV
backscatter channels together with the local incidence angle. The scenes are
radiometrically calibrated, corrected for thermal noise, and reprojected to the
NSIDC Sea-Ice Polar Stereographic North grid (EPSG:3413). In addition to SAR, the dataset contains passive-microwave and meteorological
context. AMSR2 brightness temperatures are obtained from the JAXA G-Portal at
18.7 and 36.5\,GHz in horizontal and vertical polarization. Because AMSR2 is substantially
coarser than the SAR observations, each context variable is spatially
aggregated over the corresponding SAR patch and standardized. The
primary SoD experiments use either the three Sentinel-1 channels or the
seven-channel Sentinel-1+AMSR2 configuration.
The resulting dataset contains complementary geographic coverage from the four
ice services. Figure~\ref{fig:cov_map} shows the spatial distribution of the
charted Sentinel-1 scenes. 

\begin{figure}[t]
\centering
\includegraphics[width=\columnwidth]{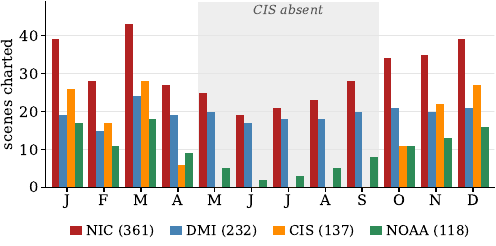}
\caption{Monthly distribution of Sentinel-1 scenes by ice service. NIC and DMI provide year-round coverage, while CIS observations are concentrated in colder months and absent from May through September.}
\label{fig:cov_month}
\end{figure}

The temporal coverage is similarly service dependent
(Figure~\ref{fig:cov_month}). In this dataset, NIC and DMI observations are represented throughout the year,
whereas CIS observations are concentrated in the colder months and are not
included from May through September. NIC
has the largest overlap with the other services, including 232 scenes with DMI,
94 with CIS, and 75 with NOAA. CIS and DMI overlap on 26 scenes, CIS and NOAA
on 57 scenes, and DMI and NOAA have no direct overlap. These co-charted scenes
provide multiple operational annotations of the same Sentinel-1 observation
and form the basis for measuring multi-annotator SoD disagreement. (One jointly
charted DMI--NIC scene shares no ice pixels and contributes no disagreement
statistics.)

The services also differ in how they express stage information. A census of the egg codes on the 404 co-charted
scenes shows that NIC delineates at the finest granularity (177 ice polygons
per scene, against 27 for DMI) yet carries the largest share of
undifferentiated stage codes (27\% of its stage entries, including code 86,
which it alone uses), whereas CIS assigns no undifferentiated codes at all and
NOAA expresses 31\% of its stage entries through the undifferentiated young-ice
code 83. DMI uses few ambiguous codes but the fullest egg-code compositions,
so its labels are wide through composition rather than code ambiguity. On
average, between 1.8 and 2.6 fine SoD classes carry nonzero proportion at an ice
pixel depending on the service. A pixel's stage label is therefore already a
distribution within a single service, before any cross-service
comparison is made.

Stage disagreement is also largely invisible to concentration-based
comparisons. Partitioning all 80.8~M co-charted pixels by whether the
services' total-concentration codes agree and whether their stage compositions
admit a common refinement, we find that 57.7\% of the pixels with agreeing
concentration codes carry provably incompatible stage compositions, with a
mean composition Wasserstein distance five times that of the fully agreeing
cell. Concentration agreement is therefore not evidence of stage agreement,
and uncertainty estimates validated against concentration disagreement have
not been validated for the SoD task considered here.

Each operational chart polygon is encoded using the WMO Egg Code and archived
in SIGRID-3 format. The Egg Code reports total ice concentration
$\mathtt{CT}$ and up to three partial ice components $j\in\{A,B,C\}$, where
each component consists of a partial concentration $C_j$ and a corresponding
stage-of-development code $S_j$. We harmonize the service-specific stage codes
onto the common SoD ladder shown in Table~\ref{tab:codes}. Specific codes map
to a single fine SoD class, whereas undifferentiated codes retain multiple
compatible stages. For example, SIGRID-3 code 86 represents undifferentiated
first-year ice and therefore supports thin, medium, and thick first-year ice.

To construct the polygon-level label, each reported partial concentration is
assigned to the class associated with its SoD code. For class $k$, the
corresponding proportion is
$\tilde{y}_{k}=\sum_{j\in\{A,B,C\}}(C_j/100)\,
\mathbf{1}[\mathrm{class}(S_j)=k]$, while the concentration not covered by sea
ice is assigned to open water as
$\tilde{y}_{\mathrm{water}}=\max(0,1-\mathtt{CT}/100)$.
Interval-valued total-concentration codes are resolved from the sum of the
reported partial concentrations when available and otherwise from the interval
midpoint. Codes 91 and 92 are treated as 100\% ice, and when only one stage is
reported without a partial concentration, that stage receives the resolved
total concentration. Water polygons are assigned directly to open water, while
bergy-water polygons are represented as 10\% glacier ice and 90\% open water.

For the ice-conditional SoD task considered in this study, open-water proportion is
removed and the remaining class proportions are renormalized over the ice
classes. Undifferentiated stage codes are retained as distributions over their
compatible fine classes, as described in
Section~\ref{sec:sod_representation}, rather than being collapsed to a single
stage.

\begin{figure*}[t]
\centering
\includegraphics[width=0.75\textwidth]{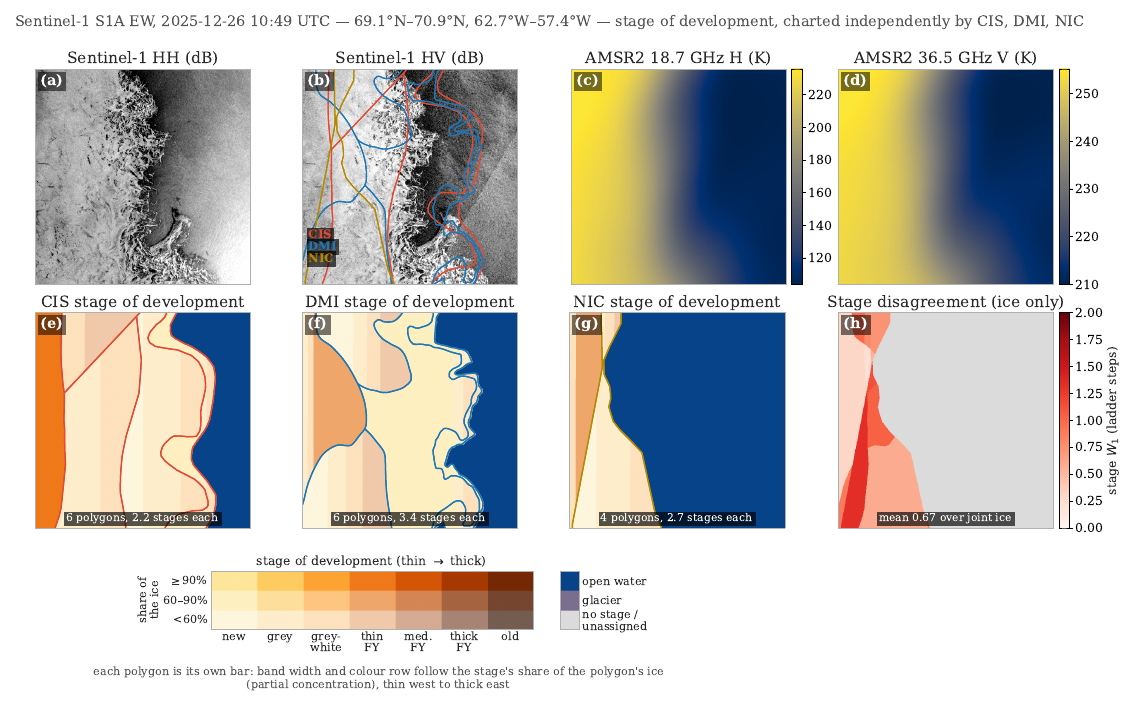}
\caption{Example of the multimodal observations and independent operational
SoD annotations for a Sentinel-1 acquisition on 26 December 2025.
(a--b) Sentinel-1 HH and HV backscatter, with the independently delineated
chart polygons overlaid in (b); (c--d) AMSR2 brightness temperatures at
18.7\,GHz H and 36.5\,GHz V polarization; (e--g) Egg-Code
stage-of-development compositions reported by CIS, DMI, and NIC; and
(h) per-pixel multi-annotator SoD disagreement measured using the first
Wasserstein distance on the ordinal stage ladder. Within each chart polygon,
the displayed stage bands summarize the reported partial-concentration shares;
their placement is a visualization of polygon composition and does not imply
spatial locations of individual ice types within the polygon.}
\label{fig:sod_scene}
\end{figure*}

Figure~\ref{fig:sod_scene} illustrates the resulting multimodal information for
one Sentinel-1 acquisition independently charted by CIS, DMI, and NIC.
Although the services observe the same SAR scene, their polygon boundaries,
partial ice compositions, and reported stages can differ. The corresponding
Wasserstein-disagreement map in Figure~\ref{fig:sod_scene}(h) makes this
variation explicit and illustrates the label uncertainty that motivates the
multi-service experiments.

Finally, the data are partitioned at the Sentinel-1 scene level. Scenes are
grouped jointly by geographic region and ISO week before splitting so that
temporally adjacent observations from the same region do not enter different
partitions. The final split contains 282 training scenes, 41 validation scenes,
and 81 held-out test scenes. All annotations associated with a Sentinel-1 scene
remain in the same split, preventing information from the same observation from
appearing in both training and evaluation.

% \begin{figure}[t]
% \centering
% \includegraphics[width=0.45\columnwidth]{figures/cov_pairs.pdf}
% \caption{Overlap of operational ice-chart coverage among the four services. Diagonal entries show total scenes per service, while off-diagonal entries show jointly charted scenes. NIC provides the main connection across regional domains; DMI and NOAA share no scenes.}
% \label{fig:cov_pair}
% \end{figure}

\subsection{Experimental Setup}

All experiments use the same single-head U-Net architecture for SoD prediction, with four encoder resolution levels containing $32$, $64$, $128$, and $256$ feature channels and an output layer producing logits over the SoD classes. Models are trained using $256\times256$ pixel crops at 80~m spatial resolution, a batch size of 16, AdamW optimization with an initial learning rate of $3\times10^{-4}$, cosine learning-rate annealing, and 30 epochs of 400 iterations each. Deep ensembles combine three independently initialized seeds for each supervision strategy, while Monte Carlo dropout applies dropout with $p=0.1$ to the decoder features feeding the output head, remaining active during both training and inference, with 16 stochastic forward passes at evaluation. Conformal prediction sets are calibrated on the validation split at $\alpha=0.1$ and evaluated on the held-out test set. To ensure consistent model selection, all models are selected using expected-thickness MAE computed against the harmonized cross-service reference distribution.

\subsection{Evaluation Metrics}
\label{sec:metrics}

We group the evaluation metrics into three categories: SoD prediction accuracy,
probabilistic quality and calibration, and uncertainty alignment with
multi-annotator disagreement.

\textbf{SoD prediction accuracy.}
We report expected-thickness MAE, coarse accuracy, and ranked probability score
(RPS). For a predicted distribution $\mathbf{p}$, expected thickness is
$\hat{T}=\sum_{k=1}^{K}p_k t_k$, where $t_k$ is the midpoint thickness of
class $k$; the reference distribution is converted similarly. MAE reports the
absolute error between these expected-thickness values in centimeters. Coarse
accuracy evaluates the modal prediction after mapping the fine SoD classes to
broader stage groups. RPS,
$\mathrm{RPS}=\sum_{k=1}^{K-1}(F_p(k)-F_q(k))^2$, evaluates the full
distribution while accounting for the ordinal distance between SoD classes.

\textbf{Probabilistic quality and calibration.}
We report negative log-likelihood (NLL), Brier score, and expected calibration
error (ECE). NLL measures the probability assigned to the reference target,
while the Brier score measures the difference between the predicted and
reference probability distributions. ECE measures the agreement between model
confidence and empirical accuracy. Lower values indicate better probabilistic
performance or calibration.

\textbf{Uncertainty alignment with multi-annotator disagreement.}
We evaluate whether model uncertainty is larger where the ice services
disagree. Predictive uncertainty is assessed using
$\rho(H,D)=\rho_{\mathrm{Spearman}}(H(x),D(x))$, where $H(x)$ is predictive
entropy and $D(x)$ is measured multi-annotator disagreement. We similarly report
$\rho_{\mathrm{epi}}=\rho_{\mathrm{Spearman}}(U_{\mathrm{epi}}(x),D(x))$ for
epistemic uncertainty. AUC$_{\mathrm{epi}}$ measures how well epistemic
uncertainty distinguishes contested from agreeing pixels, with $0.5$
corresponding to chance-level discrimination.

\subsection{Results}

\begin{table}[t]
\centering
\caption{SoD prediction and uncertainty results. MAE is expected-thickness
error in cm. $\rho(H,D)$ measures predictive-entropy alignment with
multi-annotator disagreement; $\rho_{\rm epi}$ and AUC$_{\rm epi}$ evaluate
epistemic uncertainty.}
\label{tab:main}
\setlength{\tabcolsep}{2.0pt}
\resizebox{\columnwidth}{!}{%
\begin{tabular}{lccccccccc}
\toprule
Method &
MAE &
Coarse &
RPS &
NLL &
Brier &
ECE &
$\rho(H,D)$ &
$\rho_{\rm epi}$ &
AUC$_{\rm epi}$ \\
\midrule

\multicolumn{10}{l}{\textit{SAR --- Deep ensemble}} \\

Hard
& 37.85 & 0.598 & 0.315 & 1.615 & 0.240 & 0.138 & 0.163
& -0.034 & 0.425 \\

Consensus
& \textbf{36.77} & 0.714 & \textbf{0.297}
& \textbf{1.601} & \textbf{0.234} & 0.176 & \textbf{0.193}
& 0.073 & 0.499 \\

Soft
& 40.27 & 0.620 & 0.334 & 1.614 & 0.244 & 0.106 & 0.141
& 0.013 & \textbf{0.524} \\

Weighted
& 40.84 & 0.660 & 0.338 & 1.616 & 0.245 & 0.124 & 0.184
& 0.006 & 0.496 \\

AvgNLL
& 41.47 & 0.618 & 0.359 & 1.772 & 0.273 & 0.111 & 0.139
& -0.008 & 0.495 \\

\multicolumn{10}{l}{\textit{SAR --- Consensus-target uncertainty models}} \\

MC dropout
& 38.02 & \textbf{0.719} & 0.304 & 1.616 & 0.239
& \textbf{0.051} & 0.191 & -0.049 & 0.479 \\

Evidential
& 41.43 & 0.606 & 0.401 & 1.799 & 0.279 & 0.178 & 0.090
& 0.034 & 0.425 \\

Evidential + KL cap
& 36.81 & 0.695 & 0.347 & 1.738 & 0.255 & 0.161 & 0.147
& \textbf{0.118} & 0.420 \\

\midrule

\multicolumn{10}{l}{\textit{SAR+AMSR2 --- Deep ensemble}} \\

Hard
& 39.78 & 0.626 & 0.332 & 1.609 & 0.244 & 0.132 & 0.077
& \textbf{0.182} & 0.589 \\

Consensus
& 37.68 & \textbf{0.710} & \textbf{0.288}
& 1.575 & 0.233 & 0.131 & 0.253
& 0.026 & 0.555 \\

Soft
& 41.32 & 0.580 & 0.316 & 1.569 & 0.241 & 0.128
& \textbf{0.256} & 0.023 & 0.549 \\

Weighted
& 39.30 & 0.688 & 0.306
& \textbf{1.546} & \textbf{0.229} & 0.101 & 0.220
& 0.130 & \textbf{0.608} \\

AvgNLL
& 40.50 & 0.635 & 0.335 & 1.771 & 0.262 & 0.141 & 0.207
& 0.010 & 0.512 \\

\multicolumn{10}{l}{\textit{SAR+AMSR2 --- Consensus-target uncertainty models}} \\

MC dropout
& 37.57 & 0.702 & \textbf{0.288}
& 1.565 & 0.231 & \textbf{0.050} & 0.230
& -0.121 & 0.518 \\

Evidential
& 41.93 & 0.417 & 0.434 & 1.843 & 0.295 & 0.101 & 0.125
& -0.028 & 0.400 \\

Evidential + KL cap
& \textbf{36.18} & 0.690 & 0.334
& 1.728 & 0.252 & 0.121 & 0.142
& 0.145 & 0.444 \\

\bottomrule
\end{tabular}%
}
\end{table}

Table~\ref{tab:main} compares the five supervision strategies using both SAR
and SAR+AMSR2 inputs. Preserving more annotation uncertainty does not
consistently improve SoD prediction. The deterministic cross-service consensus
is best on point and ordinal accuracy (37.68\,cm MAE, 0.710 coarse accuracy,
RPS 0.288), ahead of hard labels (39.78\,cm, 0.626). Retaining the full
cross-service distribution degrades both (41.32\,cm, 0.580), while
disagreement weighting recovers most of the loss (39.30\,cm, 0.688) and gives
the best NLL and Brier score. The uncertainty-aware targets provide only
modest gains in entropy--disagreement alignment: soft supervision gives the
highest $\rho(H,D)=0.256$, closely followed by consensus at 0.253 and weighted
supervision at 0.220.
Monte Carlo dropout on the consensus target matches consensus ordinally
(RPS 0.288, 37.57\,cm) and gives substantially lower ECE (0.050 versus
0.101--0.141 for the supervision strategies). Its predictive entropy still
correlates modestly with disagreement ($\rho(H,D)=0.230$), but its epistemic
component does not ($\rho_{\rm epi}=-0.121$,
AUROC$_{\rm epi}=0.518$). Deep-ensemble epistemic uncertainty is similarly
weak: weighted supervision gives the highest AUROC$_{\rm epi}$ at 0.608,
while most other methods remain near chance. The KL-capped evidential model
reaches $\rho_{\rm epi}=0.145$ but AUROC$_{\rm epi}=0.444$, showing that a
positive association with continuous disagreement does not necessarily yield
useful discrimination between contested and agreeing pixels.
AvgNLL tests the support-based formulation directly, rewarding mass anywhere
within the expert-supported range rather than prescribing its distribution.
The added flexibility does not help. An evidential Dirichlet head
\cite{sensoy2018evidential} on the consensus target behaves similarly: the
prescribed annealed Kullback--Leibler (KL) divergence regularization leads to
model collapse (41.93\,cm, 0.417), while capping the KL weight at 0.1 gives the
best point accuracy in the field (36.18\,cm) but a poor predictive distribution
(NLL 1.728, RPS 0.334). Both evidential variants also inflate the rare glacier
class by over an order of magnitude. Preserving only the admissible support is
therefore a weaker signal than selecting a common stage or matching the full
distribution.
Adding AMSR2 gives no consistent predictive gain: consensus moves from
36.77\,cm/0.714 with SAR alone to 37.68\,cm/0.710, AvgNLL from 41.47 to
40.50\,cm, and soft supervision from 40.27 to 41.32\,cm. Its clearest effect
is on entropy--disagreement alignment, which increases for consensus
(0.193 to 0.253) and soft supervision (0.141 to 0.256). The
brightness-temperature channels therefore add little for distinguishing SoD
classes, although they can strengthen the relationship between predictive
entropy and expert disagreement.
Errors are also structured along the SoD ladder. Thick first-year ice holds
15.5\% of the reference ice-conditioned mass but is underpredicted by every
model: soft and weighted supervision retain the most (11.3\% and 10.7\%),
followed by hard (8.4\%), consensus (6.7\%), and AvgNLL (5.2\%). Allowing
several compatible stages therefore does not resolve the first-year ambiguity
and, under AvgNLL, suppresses thick first-year ice further.

\begin{table}[t]
\centering
\caption{Split-conformal LAC sets, SAR+AMSR2, $\alpha=0.1$. $|U|$ is the
services' own answer set (mean 1.59 classes); covGap is coverage on agreeing
minus contested pixels.}
\label{tab:sod_conformal}
\setlength{\tabcolsep}{3pt}
\small
\begin{tabular}{lccccc}
\toprule
Supervision & Coverage & $|S|$ /8 & $\rho(|S|,D)$ & $\rho(|S|,|U|)$ & covGap \\
\midrule
Hard      & 0.875 & 4.02 & 0.184 & 0.306 & \textbf{0.035} \\
Consensus & 0.936 & 4.36 & \textbf{0.315} & 0.313 & 0.061 \\
Soft      & 0.910 & 4.06 & 0.302 & \textbf{0.344} & 0.095 \\
Weighted  & 0.883 & \textbf{3.83} & 0.238 & 0.311 & 0.112 \\
\bottomrule
\end{tabular}
\end{table}

Table~\ref{tab:sod_conformal} reports split-conformal LAC sets at $\alpha=0.1$. Conformal set size tracks measured disagreement in the expected direction for every supervision strategy: $\rho(|S|,D)$ is positive throughout, reaching $\rho=0.315$ for consensus, whereas $\rho_{\rm epi}$ in Table~\ref{tab:main} changes sign across methods. Predictive entropy is also positively correlated with $D$ throughout, so the contrast is specifically with the epistemic decomposition rather than predictive uncertainty as a whole. Empirical coverage is near nominal but not uniformly at or above it: hard (0.875) and weighted (0.883) fall below $1-\alpha$, potentially reflecting residual spatial dependence between calibration and test pixels.

\begin{table}[t]
\centering
\caption{Spearman correlation between predictive entropy $H(x)$ and measured
multi-annotator disagreement $D(x)$. }
\label{tab:edge_rho}
\setlength{\tabcolsep}{4pt}
\footnotesize
\begin{tabular}{lcccc}
\toprule
Supervision & 0--10 km & 10--25 km & 25--50 km & $>$50 km \\
\midrule

\multicolumn{5}{l}{\textit{SAR}} \\
Hard      & 0.523 & 0.420 & 0.281 & 0.121 \\
Consensus & \textbf{0.704} & 0.456 & 0.276 & \textbf{0.152} \\
Soft      & 0.578 & 0.469 & 0.272 & 0.068 \\
Weighted  & 0.619 & \textbf{0.491} & \textbf{0.289} & 0.120 \\
AvgNLL    & 0.174 & 0.174 & 0.124 & 0.042 \\

\midrule

\multicolumn{5}{l}{\textit{SAR+AMSR2}} \\
Hard      & 0.057 & -0.025 & 0.156 & 0.193 \\
Consensus & \textbf{0.540} & \textbf{0.309} & 0.185 & \textbf{0.248} \\
Soft      & 0.381 & 0.178 & 0.130 & 0.185 \\
Weighted  & 0.309 & 0.187 & 0.153 & 0.142 \\
AvgNLL    & 0.445 & 0.305 & \textbf{0.206} & 0.198 \\

\bottomrule
\end{tabular}
\end{table}
Table~\ref{tab:edge_rho} shows that the relationship between predictive entropy and multi-annotator disagreement is generally strongest near the ice edge. With SAR, the four supervision strategies other than AvgNLL reach 0.52--0.70 within 10\,km and decrease substantially farther into the ice pack; AvgNLL is weaker throughout (0.174 at the edge). SAR+AMSR2 shows a similar near-edge enhancement for consensus, soft, weighted, and AvgNLL, although the decline is not strictly monotonic for every method. Hard supervision is the main exception, with correlation near zero in the first two bands (0.057 and $-0.025$) and its largest value beyond 50\,km (0.193). Overall, predictive uncertainty reflects disagreement most strongly in transitional ice conditions for most supervision strategies rather than uniformly across the scene.

\begin{table}[t]
\centering
\caption{Learnability of stage disagreement. L1 is reported only for
predictors on the $W_1$ scale; distance to ice edge and predictive entropy
are rank-based scores in other units, so only $\rho$ and AUROC apply. The
constant predictor has zero variance, so $\rho$ and AUROC are undefined.}
\label{tab:sod_skyline}
\small
\begin{tabular}{lccc}
\toprule
Predictor &
$\rho(\cdot,D)$ &
AUROC &
L1 \\
\midrule
Skyline, SAR
& 0.236 & \textbf{0.574} & 0.685 \\
Skyline, SAR+AMSR2
& 0.236 & 0.490 & \textbf{0.681} \\
Distance to ice edge
& \textbf{0.384} & 0.558 & --- \\
Constant
& --- & --- & 0.712 \\
Predictive entropy
& 0.077--0.256 & 0.44--0.52 & --- \\
\bottomrule
\end{tabular}
\end{table}
Table~\ref{tab:sod_skyline} tests how predictable measured multi-annotator SoD
disagreement is from the available inputs, compared with simple reference
predictors. A weak relationship between model uncertainty and expert
disagreement could arise because the uncertainty estimator fails to capture
disagreement or because disagreement itself is difficult to predict from the
observations. The results support the latter explanation. Even when the skyline
model is trained directly to regress the measured Wasserstein disagreement, the
SAR and SAR+AMSR2 ensembles both reach only $\rho=0.236$. Their L1 errors of
0.685 and 0.681 are only slightly lower than the constant-predictor error of
0.712. Predictive entropy already reaches correlations as high as 0.256 without
being trained directly on disagreement, indicating that the disagreement signal
itself is only weakly recoverable from the available inputs.

\begin{table}[t]
\centering
\caption{Paired per-service models on shared scenes. $\bar D$ is measured
multi-annotator disagreement, $B$ is between-service model spread, $V$ is
within-service seed spread, and $B-V$ is the excess spread associated with
service-specific supervision. Coverage is the ratio of model-implied to
measured disagreement. $\Delta T_{\mathrm{model}}$ and
$\Delta T_{\mathrm{label}}$ are the signed mean expected-thickness
differences between the two services.}
\label{tab:sod_pairs}
\small
\setlength{\tabcolsep}{3.2pt}
\resizebox{\columnwidth}{!}{%
\begin{tabular}{lcccc}
\toprule
Metric &
DMI--NIC &
CIS--NIC &
NIC--NOAA &
CIS--NOAA \\
\midrule

% Scenes
% & 30 & 13 & 15 & 7 \\

Measured $\bar D$
& 1.222 & 0.575 & 1.182 & 1.130 \\

Between $B$
& 0.678 & 0.498 & 0.686 & 0.599 \\

Within $V$
& 0.307 & 0.347 & 0.470 & 0.249 \\

$B-V$
& 0.371 & 0.151 & 0.216 & 0.349 \\

$B/V$
& 2.21 & 1.44 & 1.46 & 2.41 \\

Coverage
& 0.52 & 0.70 & 0.49 & 0.50 \\

$\rho(\hat D,D)$
& 0.302 & 0.307 & $-0.055$ & 0.076 \\

AUROC
& 0.625 & 0.850 & 0.376 & 0.333 \\

$\Delta T_{\mathrm{model}}$ (cm)
& $+20.9$ & $-6.9$ & $+28.1$ & $+16.4$ \\

$\Delta T_{\mathrm{label}}$ (cm)
& $+27.1$ & $+3.5$ & $+35.8$ & $+15.2$ \\

\bottomrule
\end{tabular}%
}
\end{table}
Table~\ref{tab:sod_pairs} shows that service-specific supervision creates systematic differences in the learned models. For all service pairs, the between-service spread ($B$) exceeds the within-service seed spread ($V$), with the largest excess for DMI--NIC ($B-V=0.371$) and CIS--NOAA ($0.349$), indicating that these differences are not explained by random initialization alone. The model-implied disagreement nevertheless captures only part of the measured inter-service disagreement, as reflected by coverage values between 0.49 and 0.70. CIS--NIC provides the clearest localization of disagreement, with the highest AUROC ($0.850$) and correlation ($\rho=0.307$), while DMI--NIC shows a similar correlation ($\rho=0.302$). In contrast, NIC--NOAA and CIS--NOAA show weak or negative correlations, indicating that disagreement is harder to localize for these pairs. The signed thickness differences also generally preserve the direction of the service-specific label differences, showing that agency-specific annotation patterns leave a measurable signature in model predictions.

\begin{table}[t]
\centering
\caption{Single-service training evaluated against multi-service
disagreement. Models are trained using NIC annotations only and never observe
another ice service during training. At evaluation, predictive entropy is
compared with multi-annotator disagreement $D(x)$ computed from the unchanged
multi-service test annotations.}
\label{tab:nic_single}
\setlength{\tabcolsep}{2.0pt}
\small
\resizebox{\columnwidth}{!}{%
\begin{tabular}{lccccccc}
\toprule
Supervision &
MAE $\downarrow$ &
Coarse $\uparrow$ &
RPS $\downarrow$ &
NLL $\downarrow$ &
Brier $\downarrow$ &
ECE $\downarrow$ &
$\rho(H,D)$ $\uparrow$ \\
\midrule

\multicolumn{8}{l}{\textit{SAR}} \\

Hard
& \textbf{36.90}
& 0.602
& 0.340
& 1.716
& 0.261
& 0.124
& 0.120 \\

Soft
& 39.72
& 0.589
& \textbf{0.336}
& \textbf{1.658}
& \textbf{0.251}
& \textbf{0.104}
& \textbf{0.189} \\

AvgNLL
& 39.03
& \textbf{0.692}
& 0.351
& 1.724
& 0.261
& 0.113
& 0.185 \\

\midrule

\multicolumn{8}{l}{\textit{SAR+AMSR2}} \\

Hard
& \textbf{38.68}
& 0.642
& 0.337
& 1.693
& 0.252
& 0.135
& 0.117 \\

Soft
& 39.00
& 0.637
& \textbf{0.324}
& \textbf{1.591}
& \textbf{0.238}
& \textbf{0.097}
& \textbf{0.198} \\

AvgNLL
& 38.73
& \textbf{0.676}
& 0.342
& 1.725
& 0.254
& 0.129
& 0.132 \\

\bottomrule
\end{tabular}%
}
\end{table}
Table~\ref{tab:nic_single} shows that uncertainty--disagreement alignment persists even when the models are trained using NIC annotations alone and never observe labels from another ice service. Soft supervision provides the clearest uncertainty signal for both input configurations, giving the highest correlation between predictive entropy and inter-service disagreement for SAR ($\rho(H,D)=0.189$) and SAR+AMSR2 ($\rho(H,D)=0.198$). It also gives the best probabilistic performance, with the lowest RPS, NLL, Brier score, and ECE in both settings. Hard supervision achieves the lowest MAE, while AvgNLL gives the highest coarse accuracy, demonstrating that the method with the best point or class accuracy is not necessarily the one with the most informative or best-calibrated predictive distribution. Adding AMSR2 slightly strengthens the uncertainty--disagreement correlation for soft supervision, from 0.189 to 0.198, but does not produce a consistent improvement across the other supervision strategies. Overall, the results suggest that preserving the full within-service label distribution provides a more informative uncertainty signal than collapsing the annotation to a hard class or retaining only its supported class set.

\section{Conclusion and Limitations}
We characterized two distinct sources of uncertainty for sea-ice stage of
development and evaluated one against the other. Multi-annotator uncertainty is
measured directly from Sentinel-1 scenes charted independently by two or more
national ice services, using the first Wasserstein distance on the ordinal stage
ladder; model uncertainty is estimated from the trained networks as predictive
entropy, as an epistemic component from deep ensembles, Monte Carlo dropout, and
an evidential head, and as set-valued conformal prediction. Holding the measured
multi-annotator disagreement fixed as the reference lets us ask whether model
uncertainty recovers the disagreement among the experts, rather than treating
model confidence as evidence of annotation uncertainty on its own. Deterministic consensus provided the best predictive performance, outperforming strategies that retained more annotation information. Model uncertainty aligned most strongly with expert disagreement near the ice edge, where the SAR consensus model reached a correlation of 0.704, compared with a maximum of 0.256 across the full dataset. Monte Carlo dropout achieved the best calibration (ECE = 0.050), while conformal set size correlated consistently with disagreement, reaching 0.315. In the per-agency analysis, CIS--NIC achieved the highest disagreement discrimination (AUROC = 0.850; correlation = 0.307), while DMI--NIC showed the largest excess between-service spread ($B-V=0.371$). Across all agency pairs, between-service variation exceeded random-seed variation.

Several limitations qualify these results. They come from a single
region and season with one architecture and one calibration split,
so the ranking of supervision strategies may not transfer to other basins,
ice regimes, or split draws. Disagreement is measured across N
services, which makes $D(x)$ a coarse and possibly biased proxy for the
ambiguity an analyst perceives --- two services agreeing is weak evidence that
a pixel is unambiguous. Evaluation is also pixel-level against polygon-level
annotations, so every pixel inherits a stage assigned to a whole region, and
expected-thickness MAE depends on a fixed class-to-thickness mapping that is a
modelling choice rather than a measurement.

\begin{acks}
This research was funded by the National Science Foundation (NSF) under grant number 2531101.
\end{acks}

\bibliographystyle{ACM-Reference-Format}
\bibliography{ref}

\end{document}